\documentclass{article}

\usepackage[english]{babel}
\usepackage{amsmath}
\usepackage{amssymb}
\usepackage[letterpaper,top=2cm,bottom=2cm,left=3cm,right=3cm,marginparwidth=1.75cm]{geometry}
\usepackage{booktabs}
\usepackage{amsmath}
\usepackage{graphicx}
\usepackage{natbib}
\usepackage[colorlinks=true, allcolors=blue]{hyperref}
\usepackage{authblk}
\title{LinguaFluid: Language‑Guided Fluid Control via Semantic Rewards in Reinforcement Learning}

\author[1,2]{Aoming Liang}
\author[2]{Chi Cheng}
\author[2]{Dashuai Chen}
\author[2]{Boai Sun}
\author[2,*]{Dixia Fan}

\affil[1]{College of Environmental and Resource Sciences, Zhejiang University, Hangzhou, China}
\affil[2]{School of Engineering, Westlake University, Hangzhou, China}
\affil[*]{\texttt{fandixia@westlake.edu.cn}}

\begin{document}
\maketitle

\begin{abstract}
In the domain of scientific machine learning, designing effective reward functions remains a challenge in reinforcement learning (RL), particularly in environments where task goals are difficult to specify numerically. Reward functions in existing work are predominantly based on heuristics, manual engineering, or task-specific tuning. In this work, we introduce a semantically aligned reinforcement learning method where rewards are computed by aligning the current state with a target semantic instruction using a Sentence-Bidirectional Encoder Representations from Transformers (SBERT). Instead of relying on manually defined reward functions, the policy receives feedback based on the reward, which is a cosine similarity between the goal textual description and the statement description in the episode. We evaluated our approach in several environments and showed that semantic reward can guide learning to achieve competitive control behavior, even in the absence of hand-crafted reward functions. Our study demonstrates a correlation between the language embedding space and the conventional Euclidean space. This framework opens new horizons for aligning agent behavior with natural language goals and lays the groundwork for a more seamless integration of larger language models (LLMs) and fluid control applications.

\end{abstract}
\section{Introduction}

Reinforcement Learning (RL) has recently demonstrated remarkable potential in solving a wide range of control tasks, from robotic manipulation \citep{james2022q} and locomotion \citep{gangapurwala2022rloc}, to fluid control \citep{fan2020reinforcement,xia2024active} and the regulation of high-dimensional partial differential equation (PDE) systems \citep{alla2024online}. Despite these advances, one of the central challenges in deploying RL for scientific discovery and engineering applications lies in the design of reward functions \citep{sun2025large}. Traditional reward engineering often requires handcrafted, task-specific objective functions that may not generalize across domains, and more importantly, may fail to capture human-understandable notions of success or optimality. An interesting question arises: \emph{can a language model serve as a guiding signal for control?}. This question would provide a perspective and flexible way to encode human knowledge, enabling RL agents to align their exploration with semantically meaningful objectives. Although language models have achieved impressive success in natural language processing and have recently been coupled with RL for instruction, their application to control problems, particularly in fluid dynamics, remains unexplored. The ability to leverage semantic rewards derived from language descriptions of system states could not only reduce the need for redundant mathematical shaping of rewards but also accelerate the discovery of novel control strategies by bridging human intuition.  

Pre-trained LLMs, having been exposed to vast amounts of textual data, demonstrate a strong aptitude for language understanding and have shown great potential in high-level planning tasks such as trajectory generation \citep{xu2025trajectory,zhang2024fltrnn,lai2025llmlight}, shape optimization \citep{zhang2025using}, and simulation code generation \citep{pandey2025openfoamgpt}. However, their application to low-level control remains limited, primarily due to the inherently ambiguous and imprecise nature of their semantic outputs, which are not readily grounded in control. Collaborating with LLMs to perform specific control tasks has become a focal point of current research on how controllers can work well together. Controllers should leverage the prior knowledge and reasoning capabilities embedded in LLMs, utilizing them as cognitive modules to inform perception and decision-making \citep{ma2024reward}. Aligning controller real-time interactions with inherently static and language-based priors of LLMs poses non-trivial challenges in grounding, adaptability, and response coherence. Moreover, LLMs as agent approaches, such as function calling \citep{kim2024llm} or code-as-policy \citep{liu2024rl,liang2022code}, often suffer from significant response latency and hallucination problems \cite{ji2023survey,huang2025survey}, which is limiting for applications. There are two main reasons for this limitation. First, the nondifferentiability of frozen large language models poses a significant challenge to optimization in data-driven learning pipelines. Second, fine-tuning approaches, such as LoRA \citep{hu2022lora,cai2025dynamic}, tend to adapt models to preference alignment rather than precise dataset supervision, making them difficult to employ for direct control tasks. Most existing agent-based methods are not directly applicable to low-level control tasks and remain largely unexplored in natural scientific domains such as PDEs and active flow control. The goal of this study is to enable language models to actively participate in low-level control training energetically, allowing end-to-end guidance rather than relying on modular controller calls.

Recently, many researchers have explored heuristic approaches to adopt LLMs for downstream tasks, such as in-context learning \citep{wies2023learnability} and prompt tuning \citep{he2025dvpt}. In-context learning operates purely at inference time by conditioning the model on task-specific examples provided within the prompt, without updating any model parameters. Prompt tuning introduces a small set of trainable continuous embeddings, known as soft prompts, which are attached to the input and optimized while keeping the core LLM parameters frozen. Both approaches aim to leverage pre-trained knowledge of LLMs without complete fine-tuning, but differ significantly in terms of trainability, generalization, and stability under distribution shifts. The controller, hereafter referred to as the policy $\pi$, can be trained by reinforcement learning, as has been demonstrated in a wide range of applications. However, the design of the reward function remains a critical component in traditional RL training. 

Despite their promise, these methods remain unstable and sensitive to the behavior of LLMs. This poses a critical barrier for application, where reliable responses are required. This limitation primarily stems from the nature of LLMs: although their massive pretraining corpus enables strong capabilities in language understanding, fine-tuning is often effective for aligning input-output formats in static tasks. However, in control tasks, even a single inaccurate action can lead the system to an unrecoverable state, compounding errors, and ultimately causing failure. This sensitivity highlights the gap between LLMs and $\pi$. Since control tasks inherently involve a well-defined goal state, which can always be expressed in language, the remaining challenge lies in whether the controller can successfully navigate from the initial state to the goal state.  Our motivation is to evaluate whether LLMs can replace human-designed reward functions by aligning agent behavior with semantic goals through reward shaping. More specifically,  text such as ``The state is at $\theta = 0$'' is the goal, the dynamic description like ``The state is at $\theta = 0.5$''. The agent receives a reward proportional to the cosine similarity between these two sentence embeddings, thereby aligning its behavior with the intended semantic objective. Rather than relying on manually engineered reward functions, the agent effectively searches for high-reward trajectories by maximizing semantic similarity in the semantic space rather than Euclidean space. In this work, we demonstrate that a textual template can be generated offline using a large language model (GPT-4o \citep{hurst2024gpt}). This enables $\pi$ to implicitly learn the semantic structure encoded by the LLM, effectively establishing control in a language-based space. The advantage of using semantic representations is that they allow rapid testing \citep{qu2025latent,sun2025inverse}. However, a limitation lies in the need for prior evaluation of the semantic prompts: If the semantic distinctions are not sufficiently expressive or discriminative, the controller may not be able to explore effectively. Fortunately, the cost of such offline testing is low and can be largely delegated to large language models, which are well-suited for pre-generating or curating semantic goal spaces.

In this study, we take a first step towards this question by introducing LinguaFluid, a framework that combines semantic rewards from language models with RL agents for fluid control. Through three canonical control tasks, we demonstrate that semantic rewards are sufficiently correlated with physical metrics such as drag and can successfully guide RL agents to learn effective control strategies without the need for handcrafted numerical objectives. This opens new opportunities for language-guided scientific discovery and raises exciting future directions for designing better prompts and leveraging offline language models to provide more structured guidance in high-dimensional control settings. We demonstrate this approach in different environments, showing that agents trained with semantic rewards are capable of learning effective control strategies without access to ground-truth reward functions. Furthermore, we find that the resulting policies are often robust and interpretable, and the reward formulation is easily extensible to new goals by simply changing the instruction. The main contributions of this work are:

\begin{itemize}
    \item We define a semantic reward based on the cosine similarity between two embeddings: one derived from the natural language description of the current state, and the other from a target description of the desired state. 

    \item SBERT exhibits semantic discriminative capabilities that effectively distinguish between different states of the system, resulting in better performance in downstream control tasks.

    \item This work empirically demonstrates the successful integration of RL-LLM in fluid control. The results suggest that semantic representations can substitute for traditional elucidian space, allowing language-guided learning through the inherent exploration capabilities of RL.

\end{itemize}

\section{Methods}
\begin{figure}
    \centering
    \includegraphics[width=1\linewidth]{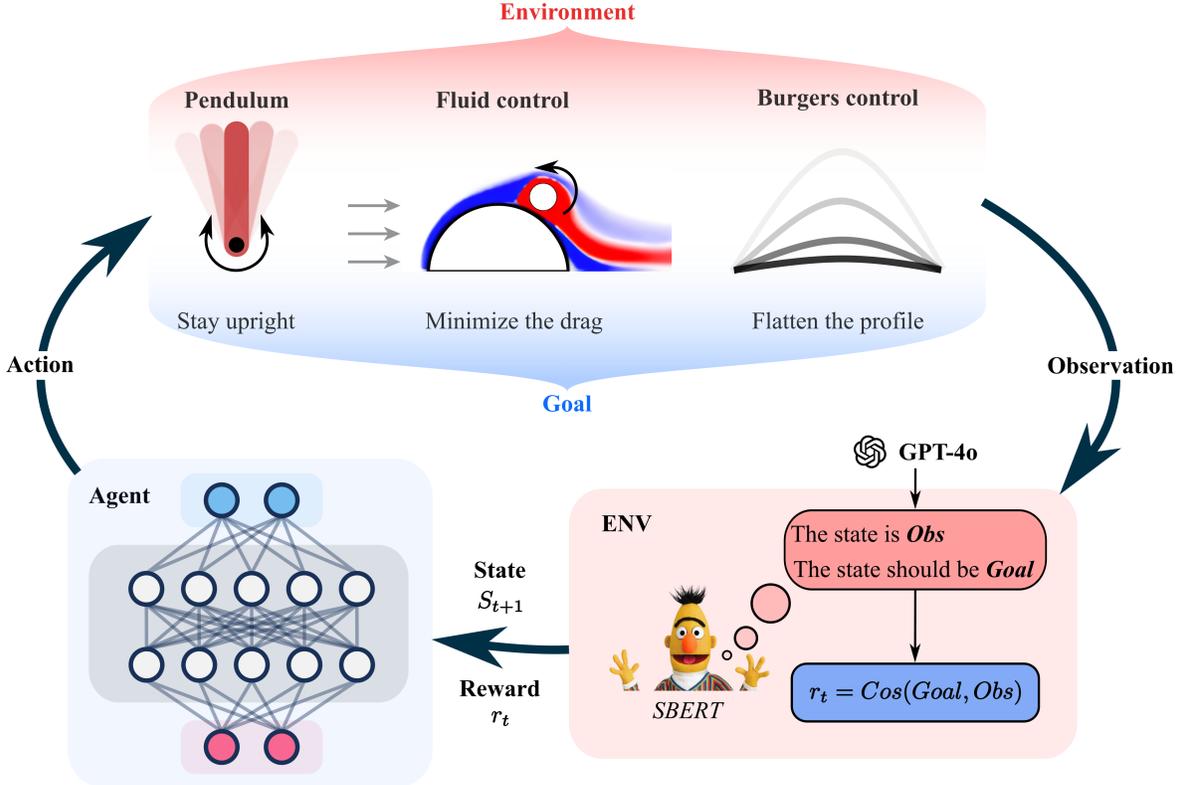}
    \caption{Semantic reward in Reinforcement learning.}
    \label{fig:fig1}
\end{figure}

The core idea of our approach is illustrated in figure~\ref{fig:fig1}. The overall pipeline follows the standard RL paradigm, with the only modification being in how the reward signal is constructed. In our setup, the language goal \( g \) is not directly provided to the policy network.  Encoder-based models, such as SBERT \citep{reimers2019sentence,chu2023refined}, generate semantically meaningful sentence-level embeddings by fine-tuning BERT using Siamese or triplet network architectures. In contrast to token-level contextual representations produced by decoder-only models such as GPT-2 \cite{radford2019language} or GPT-3 \citep{brown2020language}, SBERT directly outputs fixed-size vector embeddings that are optimized to preserve semantic similarity. This makes SBERT particularly well suited for tasks such as semantic search, clustering, and representing goal states in language-conditioned control. In this study, we adopt all‑mpnet‑base‑v2 as the language model. The choice is motivated by the fact that this model is built on the MPNet \citep{song2020mpnet} and has been further fine‑tuned on large‑scale corpora using contrastive learning, enabling it to produce high‑quality and broadly applicable sentence representations, which demonstrate strong performance in semantic matching tasks. The policy neural network ($\pi_{\theta}$) only observes the state of the environment \( s_t \) and learns the \( \pi_{\theta}(a_t \mid s_t) \) to maximize the expected cumulative reward.
\begin{equation}
J(\pi_\theta) = \mathbb{E}_{\pi_\theta} \left[ \sum_{t=0}^{T} \gamma^t r_t \right] 
= \mathbb{E}_{\pi_\theta} \left[ \sum_{t=0}^{T} \gamma^t \cos\big(\phi(g), \phi(s_t)\big) \right],
\end{equation}
where $\gamma \in [0,1)$ is the discount factor, $g$ denotes the language instruction, $s_t$ is the environment state at time $t$, and $\phi(\cdot)$ is a pretrained sentence embedding model (e.g., SBERT) that maps text into a semantic vector $ \in \mathbb{R}^{768}$. The cosine similarity $\cos(\phi(g), \phi(s_t))$ serves as a reward signal that implicitly reflects how well the agent's current state aligns with the intent of the language goal. In this work, we leverage GPT‑4o offline to generate recommended descriptions of $g$, to reduce the time required for manual template design.

Although the language goal \( g \) is not directly input into the $\pi_{\theta}$, its semantic purpose is implicitly communicated through the reward signal. For training, the agent learns to generate trajectories that align with the desired semantics of the language goal by interacting with the environment and optimizing this reward. For ease of demonstration, we show the inverted pendulum task to illustrate our approach.

\subsection{State Description via Natural Language}

The oracle pendulum environment provides state vectors in the form \( s = [\cos(\theta), \sin(\theta), \dot{\theta}] \). 
To compute semantic rewards, we first transform each state into a natural language sentence using a description textual function suggested by GPT-4o. 
\begin{equation}
\phi: \mathbb{R}^3 \rightarrow \mathcal{L},
\end{equation}
where \( \mathcal{L} \) denotes the suggestion of language descriptions through GPT-4o. 
For example, a dynamic state with \( \theta = 1.2 \) and \( \dot{\theta} = 4.0 \) is converted to the sentence:
``The state is at \( \theta = 1.20 \), \( \dot{\theta} = 4.00 \).''
 At each timestep, the reward is calculated as the cosine similarity between the embedding of the current state and the embedding of the goal.

\subsection{Proximal Policy Optimization for training}
This work adopts Proximal Policy Optimization (PPO) \citep{schulman2017proximal}, one of the most widely used RL algorithms. PPO is particularly suitable for online learning, offering greater stability, and has been widely applied in online learning tasks. We use the clipped surrogate objective from PPO:
\begin{equation}
L^{\text{CLIP}}(\theta) = \mathbb{E}_t \left[ 
\min \left( r_t(\theta) \hat{A}_t, 
\text{clip}(r_t(\theta), 1 - \epsilon, 1 + \epsilon) \hat{A}_t 
\right)
\right],
\end{equation}
where the probability ratio is:
\begin{equation}
r_t(\theta) = \frac{\pi_\theta(a_t \mid s_t)}{\pi_{\theta_{\text{old}}}(a_t \mid s_t)},
\end{equation}
and \( \hat{A}_t \) is the advantage function estimated using the generalized advantage estimation (GAE) method.

The value function V is trained to regress the semantic reward-to-go:
\begin{equation}
V(s_t) \approx \sum_{t'=t}^{T} \gamma^{t'-t} r_{t'}.
\end{equation}
A more detailed introduction is presented in the Appendix~\ref{hyper_ppo}.

\section{Results and Discussion}
To evaluate the generality and compatibility of our approach with the task, we tested it on three representative tasks: (1) stabilization of the pendulum, (2) regulation control of the Burgers equation, and (3) reduction of drag in fluid dynamics. We use the Stable-Baselines3 library for training and logging. Experiments are tracked using Weights \& Biases. In this work, we adopt the semantic prompts recommended by GPT‑4o, as shown in the Appendix~\ref{appendix:prompt}.
\subsection{Results of Pendulum}
We evaluate our method on the \texttt{Pendulum-v1} environment from OpenAI Gym \citep{brockman2016openai}, a standard control benchmark. The goal is to swing up and stabilize a single-link pendulum in an upright position. The observation is a 3-dimensional vector consisting of $\cos(\theta)$, $\sin(\theta)$, and angular velocity $\dot{\theta}$. The action is a scalar torque $a_t \in [-2, 2]$ applied to the joint. The raw reward is defined as follows:

\begin{equation}
r_t = -\left(\theta^2 + 0.1\, \dot{\theta}^2 + 0.001\, a_t^2 \right).
\end{equation}

\begin{figure}
    \centering
    \includegraphics[width=1\linewidth]{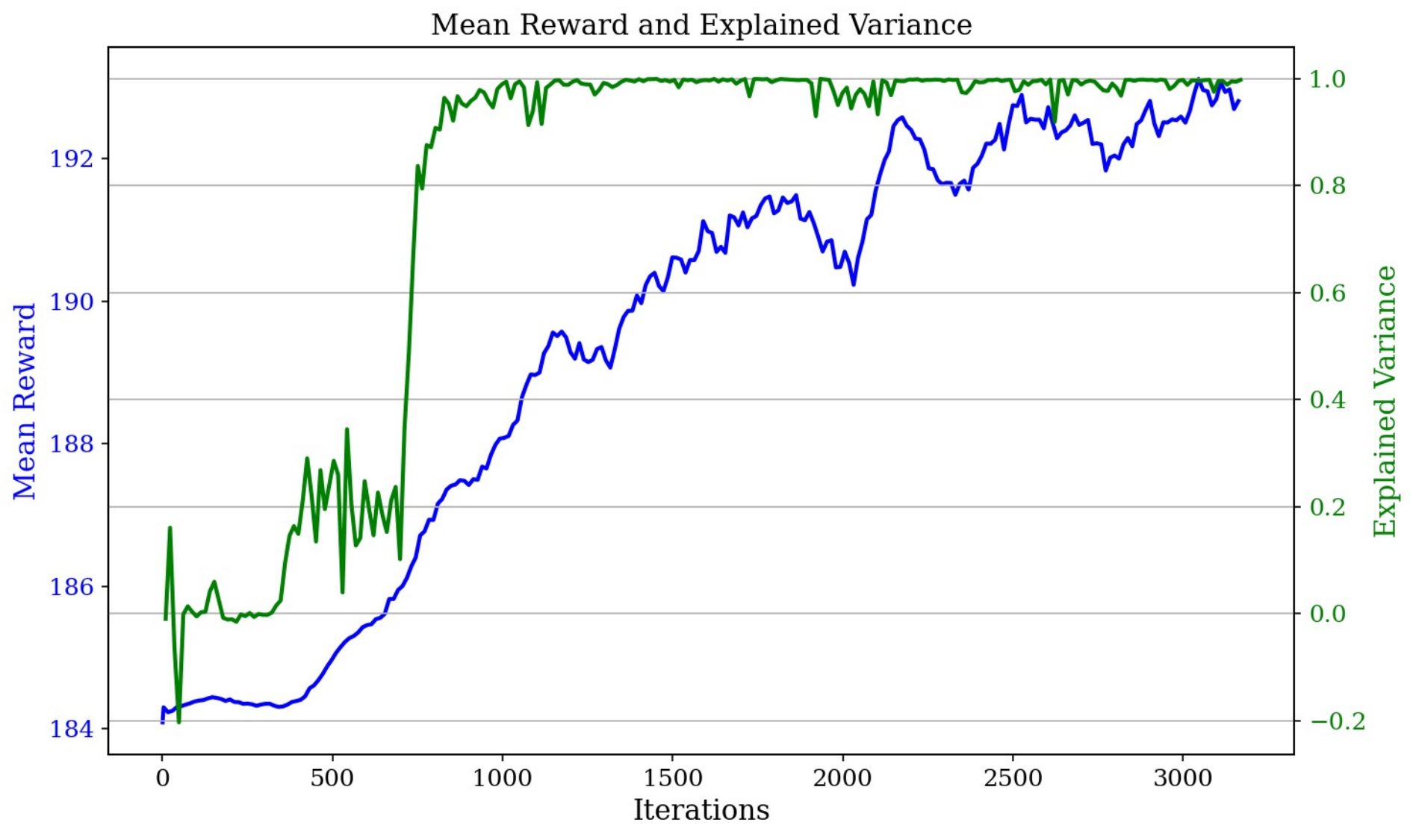}
    \caption{Evolution of Mean Reward and EV During Pendulum Training.}
    \label{fig:reward_curve_pendulum}
\end{figure}

In this work, we adopt the same environmental setup as in the original configuration. However, it is worth noting that the reward function is replaced with a language similarity-based reward for training the model. To illustrate the training process, we present the evolution of both the reward signal and the explained variance over episodes. To evaluate the quality of training progress, we report the mean reward curves and explained variance (EV) in the figure \ref{fig:reward_curve_pendulum}. The EV is defined as:

\begin{equation}
\text{EV} = 1 - \frac{\mathrm{Var}(V_{\text{target}} - V_{\text{pred}})}{\mathrm{Var}(V_{\text{target}})}.
\end{equation}

\begin{figure}
    \centering
    \includegraphics[width=1\linewidth]{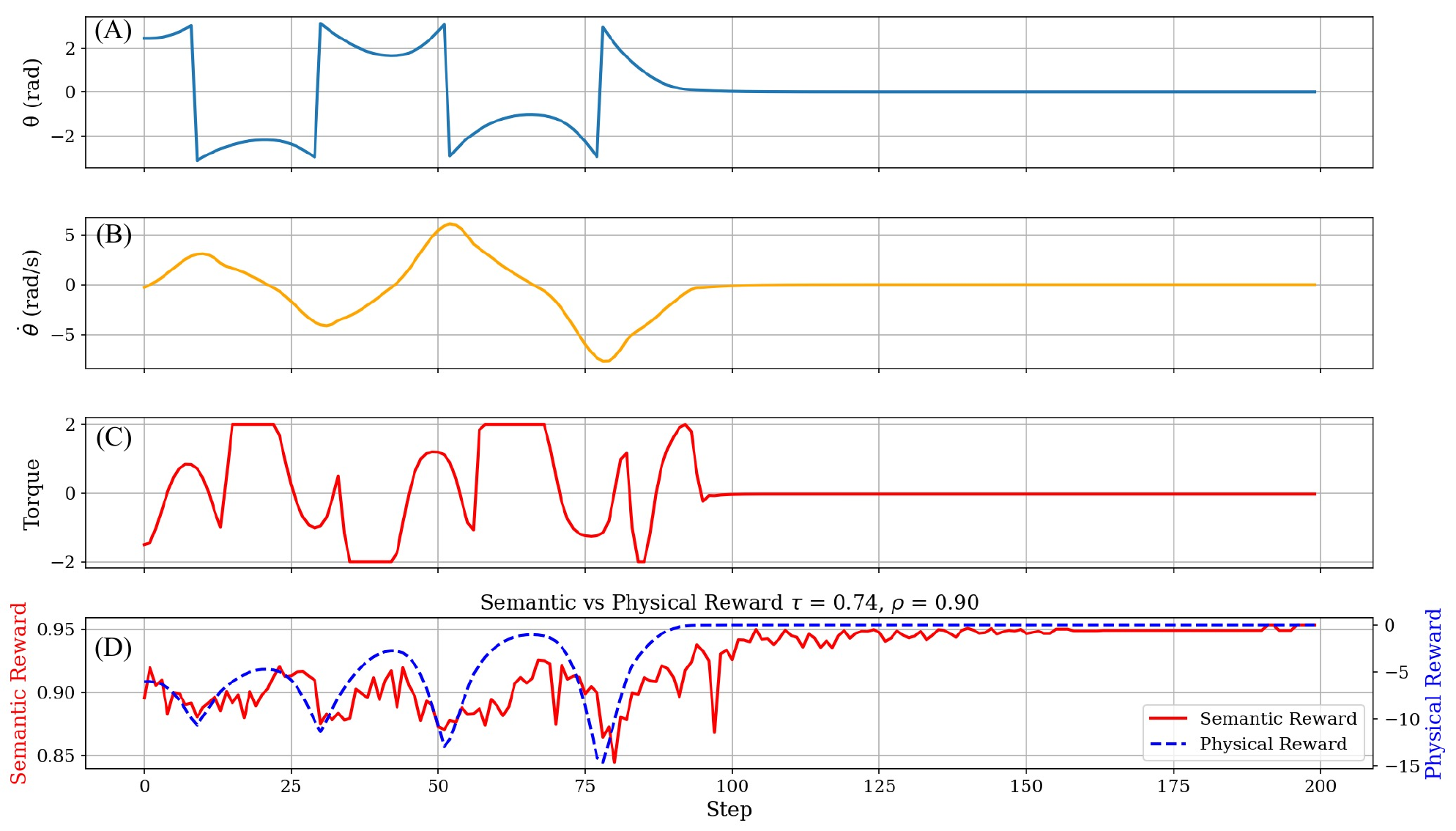}
    \caption{
    A rollout trajectory in the Pendulum environment. 
    (A) Angular position $\theta$, 
    (B) Angular velocity $\dot{\theta}$, 
    (C) Control torque, and 
    (D) Comparison between the semantic reward and the original environment reward. 
    The semantic reward is computed via SBERT similarity, while the environment reward follows the standard formulation of the Pendulum task. $\tau=0.74$ and $\rho=0.90$ indicate a strong correlation between the two reward signals, with hypothesis test $p$-values below $10^{-50}$.
    }

    \label{fig:pendulum_rollout_performance}
\end{figure}

As shown in the figure \ref{fig:reward_curve_pendulum}, the blue and green curves represent the average episode reward and the EV, respectively. Both curves increase and stabilize over training, indicating improved policy performance and more accurate value estimation in the semantic reward setting. Since the pendulum's initial state is randomly initialized, we display the result of a rollout trajectory to demonstrate the policy's behavior after training qualitatively, as shown in the figure~\ref{fig:pendulum_rollout_performance}. 
To assess the monotonic association between the semantic reward and the environment reward, Kendall's $\tau$ and Spearman's $\rho$ are defined as:
\begin{align}
\tau &= \frac{(n_c - n_d)}{\frac{1}{2}n(n-1)}, \\
\rho &= 1 - \frac{6 \sum d_i^2}{n(n^2 - 1)},
\end{align}
where $n_c$ and $n_d$ denote the number of concordant and discordant pairs, and $d_i$ is the rank difference for the sample $i$. $\tau$ focuses on the ordinal concordance between paired observations, offering a probabilistic interpretation of the consistency of the ranking, while $\rho$ measures the strength of the monotonic association based on rank correlation; both are used to assess nonparametric dependencies when linear assumptions are not appropriate. A higher value of $\tau$ and $\rho$ indicates stronger ordinal agreement.  Kendall's $\tau$ and Spearman's $\rho$ range from $-1$ to $1$. $1$ indicates a perfect positive monotonic relationship, $-1$ indicates a perfect negative monotonic relationship, and $0$ indicates no monotonic association. A broader statistical result is provided in table~\ref{tab:correlation_rewards}.

From figure~\ref{fig:pendulum_rollout_performance}A and figure~\ref{fig:pendulum_rollout_performance}B, the pendulum's angle and angular velocity are gradually corrected toward zero. Additionally, the control torque in Figure~\ref{fig:pendulum_rollout_performance}C exhibits initial oscillations. It gradually converges to zero, indicating that the pendulum stabilizes near the upright position around step 100, after which minimal energy is required to maintain balance. Figure~\ref{fig:pendulum_rollout_performance}D compares the semantic reward (red line) with the original reward (blue line), revealing a strong correlation between the two. Kendall $\tau = 0.74$ and Spearman $\rho = 0.9$, suggesting that the semantic reward in the language space is closely aligned with the original reward. This demonstrates the effectiveness of the proposed method.

\subsection{Results of Nonlinear Burgers Control}
\begin{figure}
    \centering
    \includegraphics[width=1\linewidth]{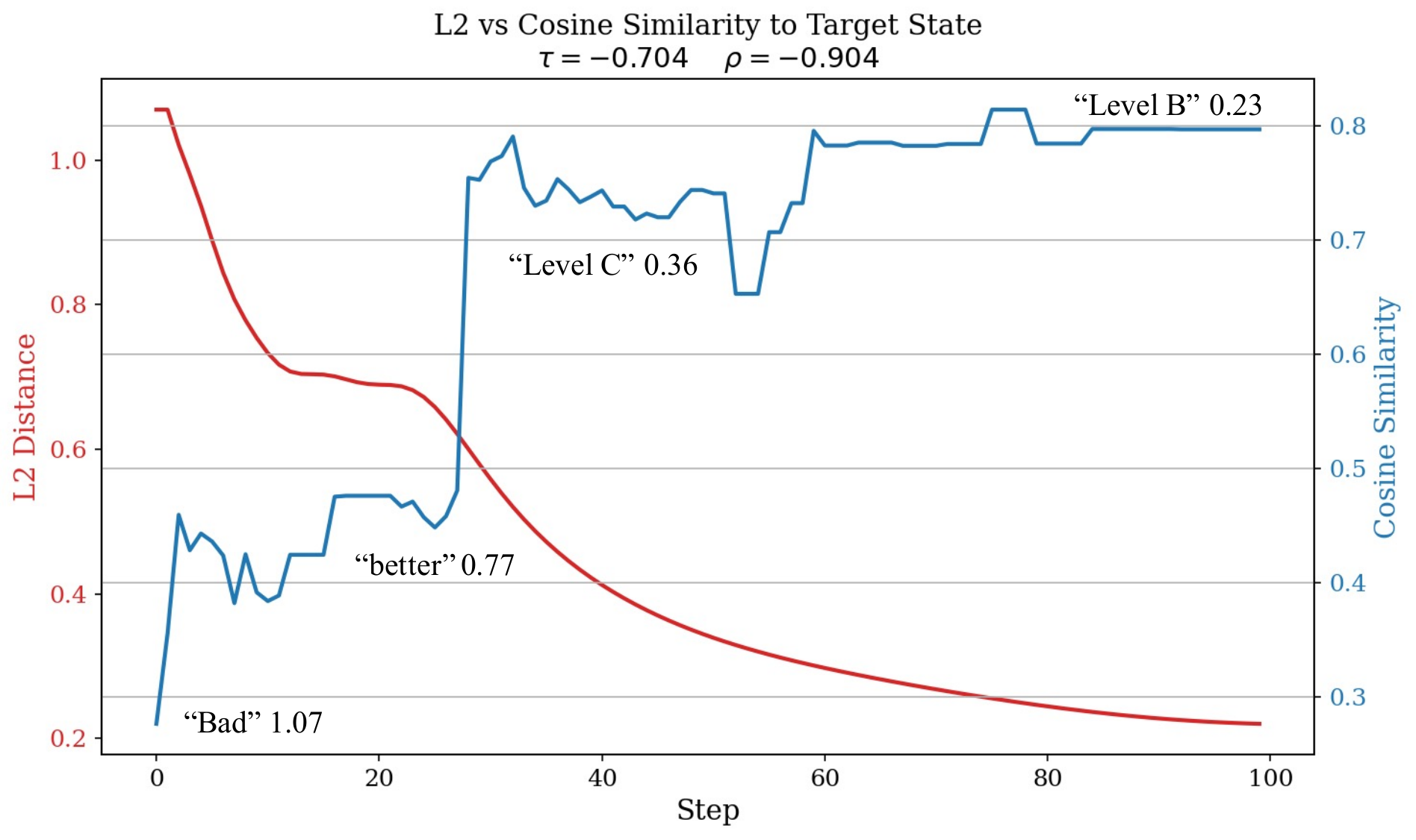}
    \caption{Evolution of the $L^2$ distance and the corresponding semantic-space similarity over time, with hypothesis test $p$-values below $10^{-20}$. Four representative descriptions are shown in the language space.}

    \label{fig:burgers_sim}
\end{figure}

The nonlinear Burgers equation with external control input is given by:
\begin{equation}
\frac{\partial u}{\partial t} + u \frac{\partial u}{\partial x} - \nu \frac{\partial^2 u}{\partial x^2} = a(x,t),
\label{eq:burgers_control}
\end{equation}
where \( u(x,t) \) denotes the velocity field, \( \nu > 0 \) is the viscosity coefficient, and \( a(x,t) \) represents the control input applied to the system. The Burgers equation serves as a canonical non-linear PDE in fluid mechanics. This equation captures key features of advective–diffusive processes, shock formation, and dissipation, making it widely used in the study of non-linear wave propagation and control in fluid flows.

\begin{equation}
a(x, t) = \sum_{i=0}^{7} a_i(t) \cdot \phi_i(x),
\end{equation}
where \( \phi_i(x) \) is a rectangular bump function defined in the reference. In this example, the viscosity is set to \(\nu = 0.001\), and the state uses 10 sparse observations.  From the figure~\ref{fig:burgers_sim}, the states are effectively controlled, with the $L^2$ distance 
$\left(\|u\|_2 = \left( \sum_{i=1}^{10} u_i^2 \right)^{1/2}\right)$  suppressed around 0.2, indicating that the energy is well regulated. Although the correlations, as measured by $\tau=0.7$ and $\rho=0.9$, are relatively high despite the sparsity, the monotonicity is not strictly increasing, which may lead to convergence to local optima. We also present the evolution of the Burgers equation in the absence of control, as shown in the Appendix~\ref{uncontrolled burgers}. From the figure~\ref{fig:burgers_reward}, while both the rewards and EV display minor oscillations, their overall stability indicates that language-driven exploration retains robustness despite inherent limitations in high-dimensional spaces. These limitations may arise from irregularities in the GPT‑4o–generated templates in the Appendix~\ref{appendix:prompt}, but the system continues to exhibit active learning capabilities.

As shown in the figure \ref{fig:burgers_with_control}A, along the temporal axis, the states gradually evolve from their initial strong fluctuations (in red) toward near-zero states (lighter colors), demonstrating that the policy successfully outputs the correct control signals. In the figure \ref{fig:burgers_with_control}B, since the total simulation horizon is 100 steps, while the action sequence spans only 99 steps, the state evolution is aligned accordingly. Finally, as illustrated in Figure \ref{fig:burgers_with_control}C, the semantic reward curve shows a consistent upward trend, further confirming the effectiveness of the learned control strategy.

\begin{figure}
    \centering
    \includegraphics[width=1\linewidth]{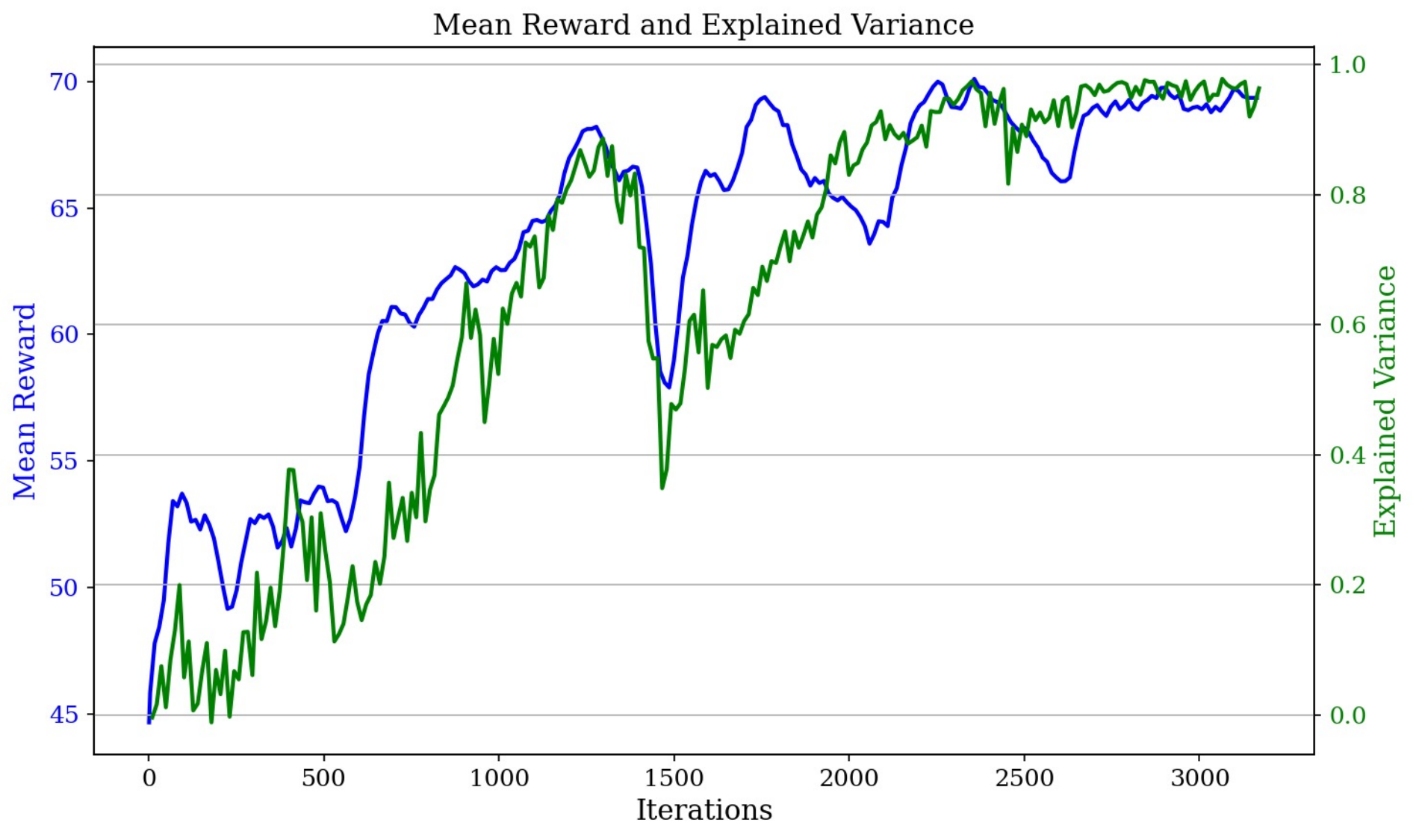}
    \caption{Evolution of Mean Reward and EV During Burgers Training.}
    \label{fig:burgers_reward}
\end{figure}
\begin{figure}
    \centering
    \includegraphics[width=1\linewidth]{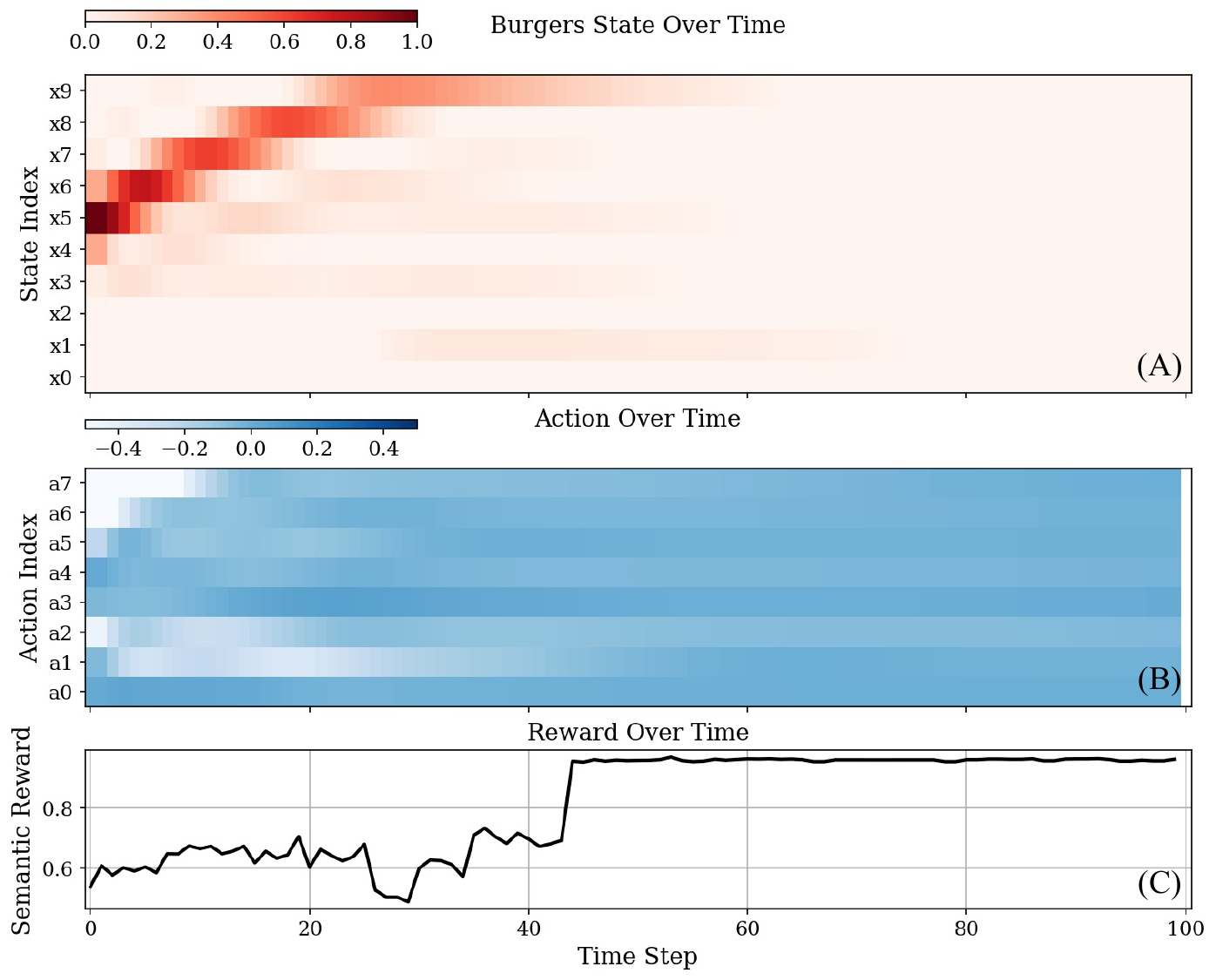}
    \caption{Evolution of the Burgers’ equation dynamics and control: (A) state evolution of the system, (B) coefficients of the right-hand-side terms of the governing equation, and (C) semantic reward trends in the language space.}
    \label{fig:burgers_with_control}
\end{figure}

\subsection{Results of drag reduction in the fluid}
In the flow around a blunt body, the width and strength of the wake directly determine the overall drag. Conventional large-scale separation leads to significant pressure differential drag. In recent years, active flow control using small-scale appendages or rotating small cylinders has emerged as an effective technique for reducing drag. This study focuses on a semi-cylinder and a rotated cylinder combination, exploring the effects of varying the rotational speed of the small cylinder on the downstream flow field. The numerical simulations are conducted using the WaterLily \citep{weymouth2025waterlily}. As illustrated in Appendix~\ref{geometry}, the computational domain consists of a large bluff body of diameter $D=48$ and a smaller downstream control cylinder. The Reynolds number is calculated as $Re=\frac{U D}{\nu}=500$, based on the dimensionless inflow velocity $U=1$, kinematic viscosity $\nu$, and the reference length $D$. The control parameter $\xi$ prescribes the angular velocity of the small cylinder according to  
\begin{equation}
    \dot{\theta}(t) = \frac{\xi U}{2d},
\end{equation}
where $d$ is the radius of the control cylinder. Since the fluid requires a certain period to develop and computational 
resources are limited, we adopt a warm-up strategy during training by 
keeping the initial $\xi = 0$ within the nondimensional time $tU/D \leq 2$. 
From $tU/D = 2.1$ to $tU/D = 3.6$, a total of 30 discrete steps are considered, 
over which the policy is trained to stabilize the propulsive power around zero, which implies the ideal condition of neither thrust nor drag.
By tuning $\xi(t)$, the rotated cylinder manipulates the wake dynamics to minimize the scaled power of the overall configuration, ideally approaching zero.  The propulsive power coefficient is defined as
\begin{equation}
C_P = \frac{F_{\mathrm{thrust}} \, U}{\rho U^3 D},
\end{equation}

\begin{figure}
    \centering
    \includegraphics[width=1\linewidth]{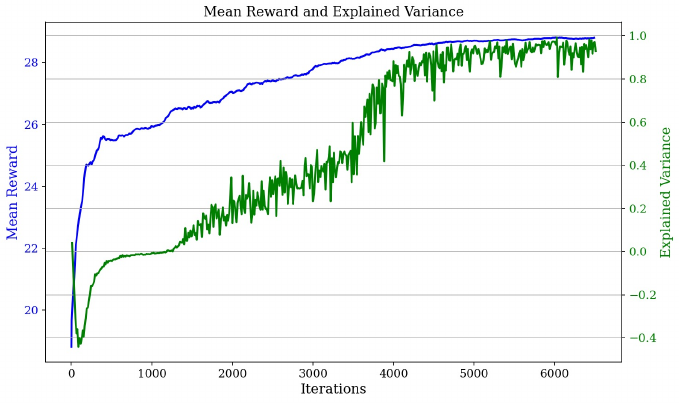}
    \caption{Evolution of Mean Reward and EV During Fluid Control Training.}
    \label{fig:fluid_reward}
\end{figure}

\begin{figure}
   \hspace*{-1cm} 
    \centering
    \includegraphics[width=1.2\linewidth]{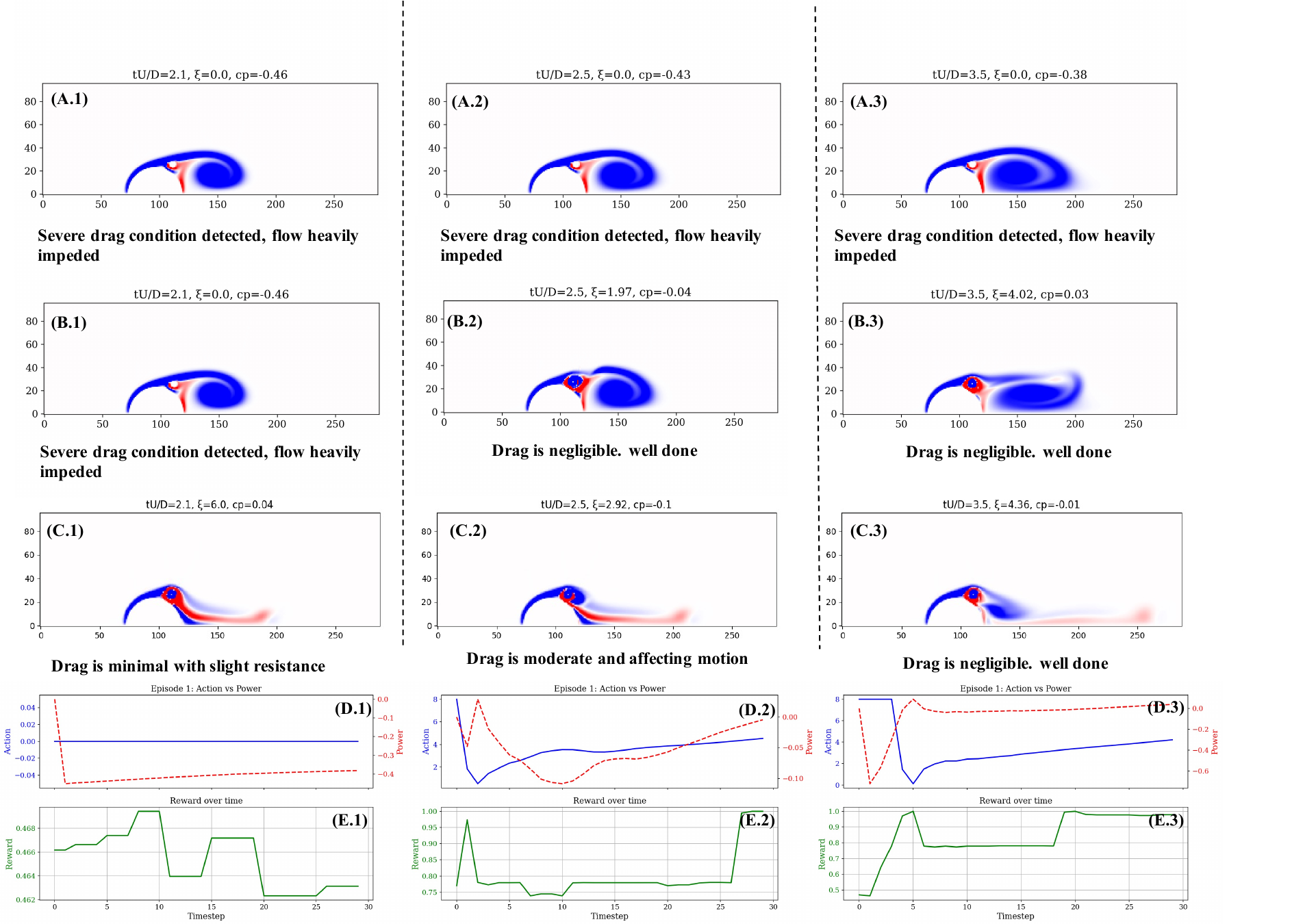}
    \caption{Comparison of system performance under different initial values of $\xi$ and corresponding state-related language descriptions.
Figures (A.1–A.3) show the uncontrolled case with $\xi = 0$ at dimensionless times $tU/D = 2.1, 2.5, 3.5$.
Figures (B.1–B.3) illustrate the controlled case with $\xi = 0$ at the same times.
Figures (C.1–C.3) present the controlled case with initial $\xi = 6$.
Figures (D.1) and (E.1) correspond to the action, drag power, and semantic rewards for case (A);
Figures (D.2) and (E.2) correspond to case (B);
and Figures (D.3) and (E.3) correspond to case (C).}
    \label{fig:drag_performance}
\end{figure}

\begin{figure}
    \centering
    \includegraphics[width=1\linewidth]{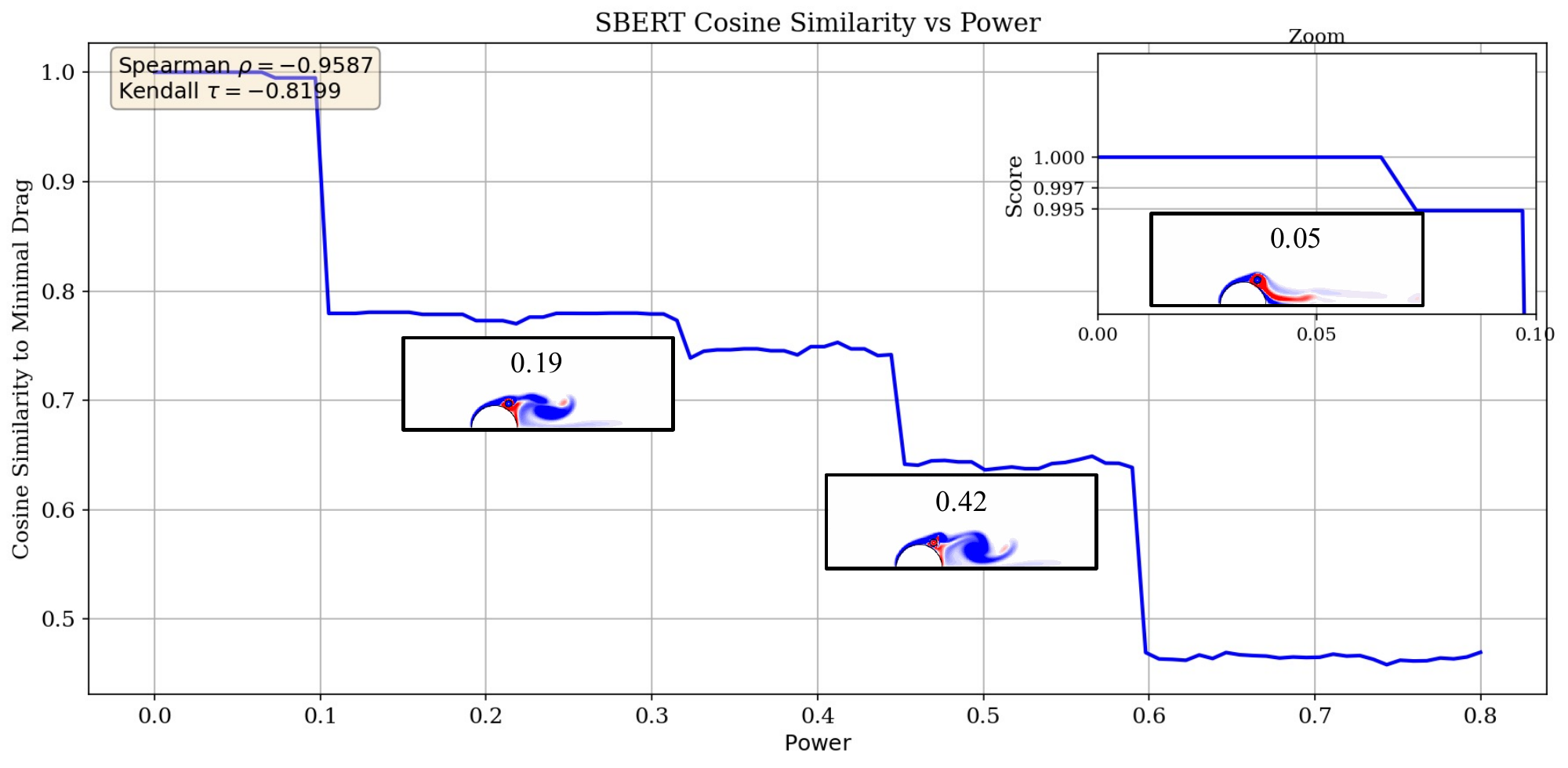}
    \caption{Relationship between power coefficient and semantic similarity, where positive values on the x-axis indicate drag.}
    \label{fig:Relationship}
\end{figure}

The evolution curves of reward and EV are shown in the figure~\ref{fig:fluid_reward}. The fast reward convergence indicates that the semantic space provides effective guidance, enabling the progressive completion of the flow control task. The comparison between the uncontrolled and controlled cases is presented in the figure~\ref{fig:drag_performance}.  In figure~\ref{fig:drag_performance}A, without control, the K\'arm\'an wake gradually develops, leading to relatively high drag power. The semantic reward remains low, and the drag power coefficient remains around 0.4. In figure~\ref{fig:drag_performance}B, with policy applied, the drag is progressively reduced. As shown in figure~\ref{fig:drag_performance}B.3, the wake structure is disrupted, the drag power coefficient remains close to zero, and the reward is significantly higher. In the unseen case with $\xi = 6$, shown in figure~\ref{fig:drag_performance}C, the policy also generalizes successfully. Since the baseline drag at $\xi = 6$ is already relatively low in figure~\ref{fig:drag_performance}C.1, the policy aims to maintain a slender wake structure and generates a slight thrust in figure~\ref{fig:drag_performance}C.2. We demonstrate the correlation between the semantic reward and the actual power coefficient, as shown in figure~\ref{fig:Relationship}. Although the curve is not perfectly smooth, a strong correlation can still be observed, with Kendall’s $\tau=-0.81$ and $\rho=-0.95$ confirming the relationship. Although local flat regions can be observed in the ranges of 
$0$--$0.05$, $0.1$--$0.3$, $0.4$--$0.55$, and $0.6$--$0.8$---indicating that 
the semantic reward is relatively insensitive in these intervals and the 
drag power exhibits limited discriminability---the results already 
demonstrate the potential of semantic rewards for reducing drag in 
fluid control. Designing more effective prompts remains an important 
direction for future work.

\begin{figure}[htbp]
    \centering
    \includegraphics[width=1\linewidth]{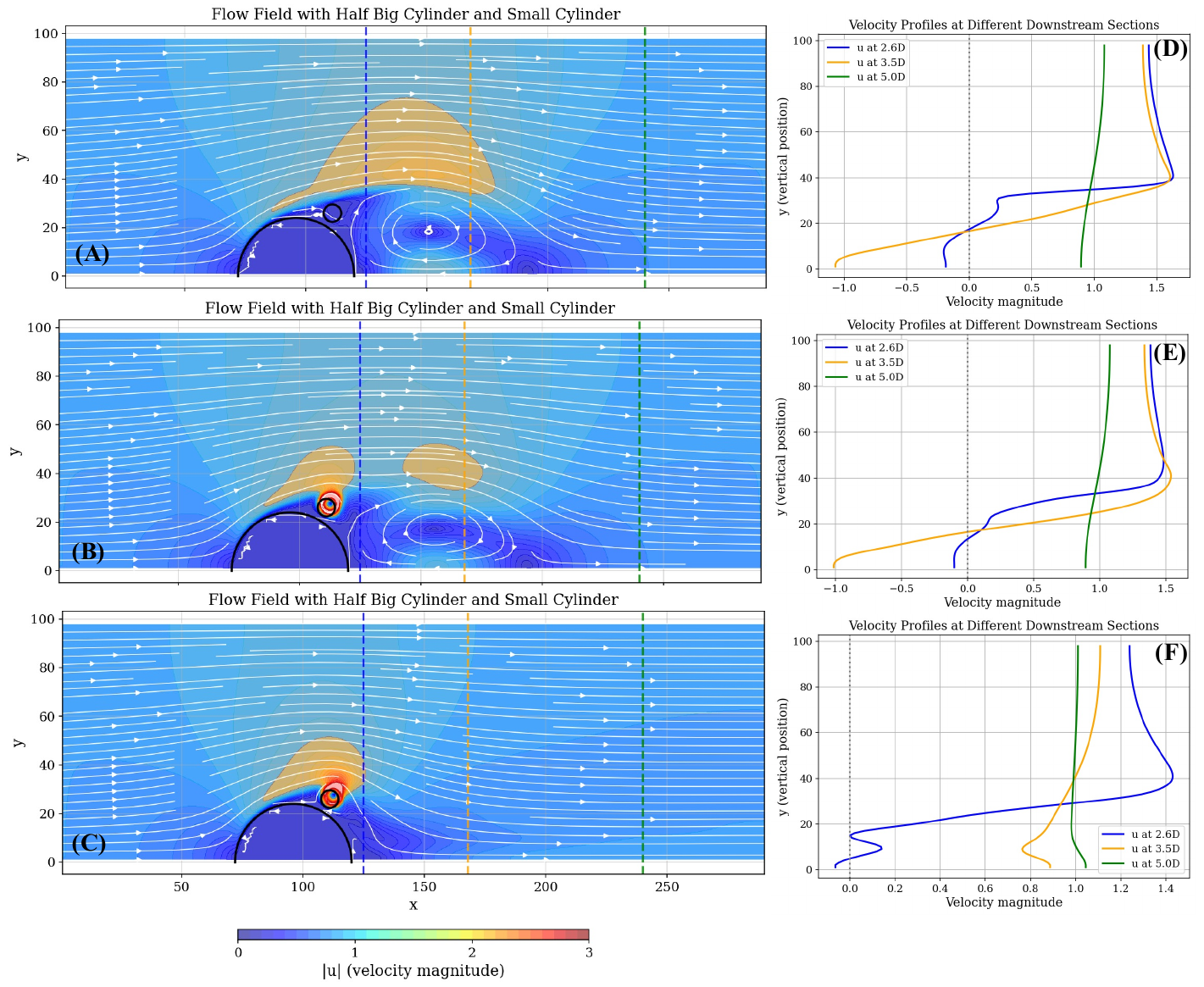}
    \caption{ Time-averaged velocity field over the simulation cases. (A) show the uncontrolled case under natural flow conditions ($\xi=0$). (B) the flow field at the onset ($\xi=0$), and (C) onset $\xi=6$, which serves as the initial condition. (D–F) present the time-averaged velocity profiles at different downstream locations, including 2.6D, 3.5D, and 5D, to describe the evolution of the velocity.}
    \label{fig:cylinder_vel}
\end{figure}

In the uncontrolled case (figure~\ref{fig:cylinder_vel}A and D), the wake behind the semi-cylindrical body exhibits a typical yellow zone of large-scale recirculation, characterized by a wide region of velocity deficit and the presence of reverse flow within the range of 2.6D to 5.0D downstream. The slow recovery of the velocity profile indicates severe momentum loss, suggesting that pressure drag dominates and the overall drag force remains relatively high. When the small cylinder is set into rotation (figure~\ref{fig:cylinder_vel}B and C), the flow field is substantially modified. The rotation induces localized vortical structures, which energize the shear layer around the small cylinder and promote partial flow reattachment on the surface of the semi-cylinder. As a result, the transverse extent of the wake is significantly reduced, as clearly observed in the streamline patterns. The corresponding downstream velocity profiles (figure~\ref{fig:cylinder_vel}E and F) demonstrate a notable attenuation of the velocity deficit, particularly at 2.6D where the reverse flow is suppressed. At further downstream locations (3.5D and 5.0D), the profiles gradually approach the free-stream condition, indicating accelerated wake recovery. The underlying mechanism of drag reduction can be attributed to the combined effects on the wake structure. On the one hand, the reduction in wake width and recirculation zone decreases the size of the low-pressure region, thereby lowering the pressure drag.  Consequently, the principal contribution of the rotating small cylinder lies in suppressing large-scale vortex shedding and mitigating velocity deficits, which leads to an overall reduction in drag force.

\subsection{Comparative Experiments on Semantic Reward}

In this section, we compare the performance of the standard PPO algorithm across three environments. For reward design, we adopt simple formulations: in Pendulum, we use the original OpenAI‑Gym setting; in Burgers, the reward is based on the negative L2 norm; and in fluid control, the reward is defined as the $r = -|C_{p}|$. The final performance is therefore evaluated concerning these reward functions. We then compare the results of our proposed language‑driven PPO with the traditional baseline PPO, as summarized in table \ref{tab:correlation_rewards}. Since both the Pendulum and Burgers involve randomness within initial conditions, we sampled 100 rollout trajectories and computed the mean and standard deviation of the rewards.

\begin{table}[htbp]
\centering
\begin{tabular}{|c|c|c|c|c|}
\hline
\textbf{Task} & \textbf{Mean of Kendall's \( \tau \)} & \textbf{Mean of Spearman's \( \rho \)} & \textbf{R. Reward} & \textbf{T. PPO Reward}\\
\hline
Pendulum & -0.60 & -0.84 & -129 $\pm$164& -178$\pm$30\\
Burgers  & -0.61 & -0.79 & -42 $\pm$3.4 & -31$\pm$2.6   \\
Fluid control & -0.82 & -0.95 & -38 & -35 \\
\hline
\end{tabular}
\caption{Correlation and reward comparison across tasks. R. Reward means Semantic-based PPO on the task. T. PPO means traditional PPO without semantic rewards.}
\label{tab:correlation_rewards}
\end{table}
From table~\ref{tab:correlation_rewards}, it can be observed that across the three tasks, the performance of the proposed semantic‑based PPO is nearly identical to that of the standard PPO. The average Kendall and Spearman correlations remain relatively high. In the Pendulum task, the mean reward reaches around 129, although with a relatively large variance due to the sensitivity to initial conditions. In the Burgers control task, the performance is approximately 10 points lower than that of the baseline PPO, which can be attributed to the inherent sparsity of the semantic space, particularly in high-dimensional state-action settings where monotonicity is not strictly preserved. In the fluid control task, the results are essentially the same as those obtained with the standard method, indicating that language‑driven control is feasible for low‑dimensional and nonlinear tasks.

\newpage
\section{Conclusion}
In this study, we investigated the integration of RL-fluid utilizing semantic rewards in LLMs. Across three representative control tasks, we demonstrate that semantic rewards are not only feasible but also effective in guiding agents toward meaningful behaviors. Importantly, this approach eliminates the need for carefully engineered reward functions, offering a more flexible pathway for controlling complex dynamical systems such as fluid flows. Despite the inherently fuzzy nature of natural language, we showed that semantic guidance can successfully shape the exploration process. Our analysis further revealed a strong correlation between the rewards defined in the semantic space and the physical measures in the Eulerian space, highlighting the potential of language as a high-level supervisory signal. However, several limitations remain. The current framework relies on relatively simple prompt templates and task descriptions, which may restrict its generalization to broader physical scenarios. Moreover, while semantic rewards were found to correlate well with drag and power coefficients, mapping from natural language to physical actions is still indirect and could be sensitive to prompt design. For high-dimensional tasks such as the Burgers equation, the performance is still not particularly satisfactory, primarily because the rewards in the language space are not strictly monotonic. This limitation, which cannot be fully addressed by the current reliance on language templates, calls for further in-depth exploration in future work.

\bibliographystyle{unsrt}
\bibliography{main}

\begin{thebibliography}{10}

\bibitem{james2022q}
Stephen James and Andrew~J Davison.
\newblock Q-attention: Enabling efficient learning for vision-based robotic manipulation.
\newblock {\em IEEE Robotics and Automation Letters}, 7(2):1612--1619, 2022.

\bibitem{gangapurwala2022rloc}
Siddhant Gangapurwala, Mathieu Geisert, Romeo Orsolino, Maurice Fallon, and Ioannis Havoutis.
\newblock Rloc: Terrain-aware legged locomotion using reinforcement learning and optimal control.
\newblock {\em IEEE Transactions on Robotics}, 38(5):2908--2927, 2022.

\bibitem{fan2020reinforcement}
Dixia Fan, Liu Yang, Zhicheng Wang, Michael~S Triantafyllou, and George~Em Karniadakis.
\newblock Reinforcement learning for bluff body active flow control in experiments and simulations.
\newblock {\em Proceedings of the National Academy of Sciences}, 117(42):26091--26098, 2020.

\bibitem{xia2024active}
Chengwei Xia, Junjie Zhang, Eric~C Kerrigan, and Georgios Rigas.
\newblock Active flow control for bluff body drag reduction using reinforcement learning with partial measurements.
\newblock {\em Journal of Fluid Mechanics}, 981:A17, 2024.

\bibitem{alla2024online}
Alessandro Alla, Agnese Pacifico, Michele Palladino, and Andrea Pesare.
\newblock Online identification and control of pdes via reinforcement learning methods.
\newblock {\em Advances in Computational Mathematics}, 50(4):85, 2024.

\bibitem{sun2025large}
Shengjie Sun, Runze Liu, Jiafei Lyu, Jing-Wen Yang, Liangpeng Zhang, and Xiu Li.
\newblock A large language model-driven reward design framework via dynamic feedback for reinforcement learning.
\newblock {\em Knowledge-Based Systems}, page 114065, 2025.

\bibitem{xu2025trajectory}
Yi~Xu, Ruining Yang, Yitian Zhang, Yizhou Wang, Jianglin Lu, Mingyuan Zhang, Lili Su, and Yun Fu.
\newblock Trajectory prediction meets large language models: A survey.
\newblock {\em arXiv preprint arXiv:2506.03408}, 2025.

\bibitem{zhang2024fltrnn}
Jiatao Zhang, Lanling Tang, Yufan Song, Qiwei Meng, Haofu Qian, Jun Shao, Wei Song, Shiqiang Zhu, and Jason Gu.
\newblock Fltrnn: Faithful long-horizon task planning for robotics with large language models.
\newblock In {\em 2024 IEEE International Conference on Robotics and Automation (ICRA)}, pages 6680--6686. IEEE, 2024.

\bibitem{lai2025llmlight}
Siqi Lai, Zhao Xu, Weijia Zhang, Hao Liu, and Hui Xiong.
\newblock Llmlight: Large language models as traffic signal control agents.
\newblock In {\em Proceedings of the 31st ACM SIGKDD Conference on Knowledge Discovery and Data Mining V. 1}, pages 2335--2346, 2025.

\bibitem{zhang2025using}
Xinxin Zhang, Zhuoqun Xu, Guangpu Zhu, Chien Ming~Jonathan Tay, Yongdong Cui, Boo~Cheong Khoo, and Lailai Zhu.
\newblock Using large language models for parametric shape optimization.
\newblock {\em Physics of Fluids}, 37(8), 2025.

\bibitem{pandey2025openfoamgpt}
Sandeep Pandey, Ran Xu, Wenkang Wang, and Xu~Chu.
\newblock Openfoamgpt: A retrieval-augmented large language model (llm) agent for openfoam-based computational fluid dynamics.
\newblock {\em Physics of Fluids}, 37(3), 2025.

\bibitem{ma2024reward}
Haozhe Ma, Kuankuan Sima, Thanh~Vinh Vo, Di~Fu, and Tze-Yun Leong.
\newblock Reward shaping for reinforcement learning with an assistant reward agent.
\newblock In {\em Forty-first international conference on machine learning}, 2024.

\bibitem{kim2024llm}
Sehoon Kim, Suhong Moon, Ryan Tabrizi, Nicholas Lee, Michael~W Mahoney, Kurt Keutzer, and Amir Gholami.
\newblock An llm compiler for parallel function calling.
\newblock In {\em Forty-first International Conference on Machine Learning}, 2024.

\bibitem{liu2024rl}
Shaoteng Liu, Haoqi Yuan, Minda Hu, Yanwei Li, Yukang Chen, Shu Liu, Zongqing Lu, and Jiaya Jia.
\newblock Rl-gpt: Integrating reinforcement learning and code-as-policy.
\newblock {\em Advances in Neural Information Processing Systems}, 37:28430--28459, 2024.

\bibitem{liang2022code}
Jacky Liang, Wenlong Huang, Fei Xia, Peng Xu, Karol Hausman, Brian Ichter, Pete Florence, and Andy Zeng.
\newblock Code as policies: Language model programs for embodied control.
\newblock {\em arXiv preprint arXiv:2209.07753}, 2022.

\bibitem{ji2023survey}
Ziwei Ji, Nayeon Lee, Rita Frieske, Tiezheng Yu, Dan Su, Yan Xu, Etsuko Ishii, Ye~Jin Bang, Andrea Madotto, and Pascale Fung.
\newblock Survey of hallucination in natural language generation.
\newblock {\em ACM computing surveys}, 55(12):1--38, 2023.

\bibitem{huang2025survey}
Lei Huang, Weijiang Yu, Weitao Ma, Weihong Zhong, Zhangyin Feng, Haotian Wang, Qianglong Chen, Weihua Peng, Xiaocheng Feng, Bing Qin, et~al.
\newblock A survey on hallucination in large language models: Principles, taxonomy, challenges, and open questions.
\newblock {\em ACM Transactions on Information Systems}, 43(2):1--55, 2025.

\bibitem{hu2022lora}
Edward~J Hu, Yelong Shen, Phillip Wallis, Zeyuan Allen-Zhu, Yuanzhi Li, Shean Wang, Lu~Wang, Weizhu Chen, et~al.
\newblock Lora: Low-rank adaptation of large language models.
\newblock {\em ICLR}, 1(2):3, 2022.

\bibitem{cai2025dynamic}
Guohui Cai, Anda Kai, and Fan Guo.
\newblock Dynamic and low-rank fine-tuning of large language models for robust few-shot learning.
\newblock {\em Transactions on Computational and Scientific Methods}, 5(4), 2025.

\bibitem{wies2023learnability}
Noam Wies, Yoav Levine, and Amnon Shashua.
\newblock The learnability of in-context learning.
\newblock {\em Advances in Neural Information Processing Systems}, 36:36637--36651, 2023.

\bibitem{he2025dvpt}
Along He, Yanlin Wu, Zhihong Wang, Tao Li, and Huazhu Fu.
\newblock Dvpt: Dynamic visual prompt tuning of large pre-trained models for medical image analysis.
\newblock {\em Neural Networks}, 185:107168, 2025.

\bibitem{hurst2024gpt}
Aaron Hurst, Adam Lerer, Adam~P Goucher, Adam Perelman, Aditya Ramesh, Aidan Clark, AJ~Ostrow, Akila Welihinda, Alan Hayes, Alec Radford, et~al.
\newblock Gpt-4o system card.
\newblock {\em arXiv preprint arXiv:2410.21276}, 2024.

\bibitem{qu2025latent}
Yun Qu, Yuhang Jiang, Boyuan Wang, Yixiu Mao, Cheems Wang, Chang Liu, and Xiangyang Ji.
\newblock Latent reward: Llm-empowered credit assignment in episodic reinforcement learning.
\newblock In {\em Proceedings of the AAAI Conference on Artificial Intelligence}, volume~39, pages 20095--20103, 2025.

\bibitem{sun2025inverse}
Hao Sun and Mihaela van~der Schaar.
\newblock Inverse reinforcement learning meets large language model post-training: Basics, advances, and opportunities.
\newblock {\em arXiv preprint arXiv:2507.13158}, 2025.

\bibitem{reimers2019sentence}
Nils Reimers and Iryna Gurevych.
\newblock Sentence-bert: Sentence embeddings using siamese bert-networks.
\newblock {\em arXiv preprint arXiv:1908.10084}, 2019.

\bibitem{chu2023refined}
Yonghe Chu, Heling Cao, Yufeng Diao, and Hongfei Lin.
\newblock Refined sbert: Representing sentence bert in manifold space.
\newblock {\em Neurocomputing}, 555:126453, 2023.

\bibitem{radford2019language}
Alec Radford, Jeffrey Wu, Rewon Child, David Luan, Dario Amodei, Ilya Sutskever, et~al.
\newblock Language models are unsupervised multitask learners.
\newblock {\em OpenAI blog}, 1(8):9, 2019.

\bibitem{brown2020language}
Tom Brown, Benjamin Mann, Nick Ryder, Melanie Subbiah, Jared~D Kaplan, Prafulla Dhariwal, Arvind Neelakantan, Pranav Shyam, Girish Sastry, Amanda Askell, et~al.
\newblock Language models are few-shot learners.
\newblock {\em Advances in neural information processing systems}, 33:1877--1901, 2020.

\bibitem{song2020mpnet}
Kaitao Song, Xu~Tan, Tao Qin, Jianfeng Lu, and Tie-Yan Liu.
\newblock Mpnet: Masked and permuted pre-training for language understanding.
\newblock {\em Advances in neural information processing systems}, 33:16857--16867, 2020.

\bibitem{schulman2017proximal}
John Schulman, Filip Wolski, Prafulla Dhariwal, Alec Radford, and Oleg Klimov.
\newblock Proximal policy optimization algorithms.
\newblock {\em arXiv preprint arXiv:1707.06347}, 2017.

\bibitem{brockman2016openai}
Greg Brockman, Vicki Cheung, Ludwig Pettersson, Jonas Schneider, John Schulman, Jie Tang, and Wojciech Zaremba.
\newblock Openai gym.
\newblock {\em arXiv preprint arXiv:1606.01540}, 2016.

\bibitem{weymouth2025waterlily}
Gabriel~D Weymouth and Bernat Font.
\newblock Waterlily. jl: A differentiable and backend-agnostic julia solver for incompressible viscous flow around dynamic bodies.
\newblock {\em Computer Physics Communications}, page 109748, 2025.

\end{thebibliography}

\section*{Acknowledgements}
The authors declare that this research was conducted without any financial support or funding from public, commercial, or not-for-profit sectors.

\section{Appendix}

\subsection{The semantic prompt used in the tasks}
\label{appendix:prompt}
We design semantic state describers that translate raw numerical observations into natural language prompts, which are then used to guide reinforcement learning agents. Table~\ref{tab:semantic_descriptions} summarizes the state description strategies at different levels. In addition to language-based descriptions, we also incorporate numerical values. Without numerical information, the descriptions would be too ambiguous, while including too many digits would make the embeddings overly sensitive in the language model. Therefore, we retain numerical values up to two decimal places.

\begin{table}[htbp]
\centering
\caption{Semantic state descriptions across different tasks. Numerical states are mapped into interpretable natural language prompts.}
\label{tab:semantic_descriptions}
\begin{tabular}{p{0.18\linewidth} p{0.25\linewidth} p{0.50\linewidth}}
\toprule
\textbf{Task} & \textbf{Numerical Feature} & \textbf{Semantic Description} \\
\midrule
Burgers’ Equation &
$L^2 < 0.2$ & Level A $L^2$\\
& $0.2 \leq L^2 < 0.3$ & Level B $L^2$\\
& $0.3 \leq L^2 < 0.4$ & Level C $L^2$\\
& $0.4 \leq L^2 < 0.5$ & Level D $L^2$\\
& $0.5 \leq L^2 < 0.6$ & Level E $L^2$\\
& $0.6 \leq L^2 < 0.7$ & excellent $L^2$\\
& $0.7 \leq L^2 < 0.8$ & better $L^2$\\
& $0.8 \leq L^2 < 0.9$ & good $L^2$\\
& $0.9 \leq L^2 < 1.0$ & normal $L^2$\\
& $1.0 \leq L^2 < 1.1$ & bad $L^2$\\
& $L^2 \geq 1.1$ & collapse $L^2$\\
\midrule
Pendulum &
$\arctan2(\text{obs}[1], \text{obs}[0])$, \quad $\dot{\theta} = \text{obs}[2]$ &
State expressed as \texttt{[angle = $\theta$, theta\_dot = $\dot{\theta}$]} \\
\midrule
Fluid Control &
$Cp^{2} < 0.01$ & Drag is negligible, well done \texttt{$Cp^{2}$}\\
& $0.01 \leq Cp^{2} < 0.05$ & Drag is minimal with slight resistance \texttt{$Cp^{2}$}\\
& $0.05 \leq Cp^{2} < 0.10$ & Drag is mild but noticeable \texttt{$Cp^{2}$}\\
& $0.10 \leq Cp^{2} < 0.20$ & Drag is moderate and affecting motion \texttt{$Cp^{2}$}\\
& $0.20 \leq Cp^{2} < 0.35$ & Drag is strong and significantly slowing flow \texttt{$Cp^{2}$}\\
& $Cp^{2} \geq 0.35$ & Severe drag condition detected, flow heavily impeded \texttt{$Cp^{2}$}\\
\bottomrule
\end{tabular}
\end{table}

\subsection{The hyper-parameters of PPO in the tasks}
\label{hyper_ppo}
\begin{table}[htbp]
\centering
\caption{Hyperparameters used for PPO training}
\label{tab:ppo_hyperparams}
\begin{tabular}{ll}
\toprule
\textbf{Hyperparameter} & \textbf{Value} \\
\midrule
Optimizer                       & Adam \\
Learning rate                   & $3 \times 10^{-4}$ \\
Discount factor $\gamma$        & 0.99 \\
GAE parameter $\lambda$         & 0.95 \\
Clip ratio $\epsilon$           & 0.2 \\
Policy network architecture     & 2 layers, 64 units per layer (MLP) \\
Value network architecture      & 2 layers, 64 units per layer (MLP) \\
Number of epochs per update     & 10 \\
Mini-batch size                 & 64 \\
Rollout length (steps per update) & 2048 \\
Entropy coefficient             & 0.01 \\
Value loss coefficient          & 0.5 \\
Max gradient norm               & 0.5 \\
Total training timesteps        & $1 \times 10^{6}$ \\
\bottomrule
\end{tabular}
\end{table}

\subsection{Existence of Optimal State Representations}

\textbf{Proposition} There exists a state \( s^* \in \mathcal{S} \) such that:

\[
s^* = \arg\max_{s \in \mathcal{S}} \cos\left( f(\phi(s)), f(g) \right)
\]

Assuming the embedding of the sentence \( f \) reflects semantic closeness, and the goal \( g \) is linguistically well-defined (e.g., ``keep the pendulum upright''), we expect the corresponding physical state \( s^* \) to be consistent with the intended result (e.g., \( \theta = 0, \dot{\theta} = 0 \)).

\subsection{Semantic Consistency and Gradient Structure}

\textbf{Proposition} SBERT satisfies semantic consistency, meaning that if:

\[
\phi(s_1) \approx \phi(s_2) \quad \Rightarrow \quad \cos\left( f(\phi(s_1)), f(\phi(s_2)) \right) \approx 1
\]

Therefore, the reward function \( r(s) \) is Lipschitz continuous in semantic space. That is, there exists \( L > 0 \) such that:

\[
|r(s_1) - r(s_2)| \leq L \cdot \| f(\phi(s_1)) - f(\phi(s_2)) \|_2
\]

This ensures a smooth reward landscape that can be optimized using policy gradient methods.
\subsection{Proximal Policy Optimization (PPO)}

Proximal Policy Optimization (PPO) is a widely used reinforcement learning algorithm that achieves stable and efficient policy updates by constraining the deviation from the previous policy during optimization. It is an on-policy actor-critic method that alternates between sampling trajectories using the current policy and updating the policy via gradient ascent.

\subsubsection{Objective Function}

The central idea of PPO is to optimize a clipped surrogate objective to prevent overly large policy updates. Let \( \theta \) be the parameters of the current policy, and \( \theta_{\text{old}} \) be the parameters of the policy used to collect trajectories. The probability ratio is defined as:

\[
r_t(\theta) = \frac{\pi_\theta(a_t \mid s_t)}{\pi_{\theta_{\text{old}}}(a_t \mid s_t)}
\]

The PPO objective is:

\[
L^{\text{CLIP}}(\theta) = \mathbb{E}_t \left[ \min \left( r_t(\theta) \hat{A}_t,\ \text{clip}\left(r_t(\theta), 1 - \epsilon, 1 + \epsilon\right) \hat{A}_t \right) \right]
\]

Here, \( \hat{A}_t \) is the estimated advantage function, and \( \epsilon \) is a hyperparameter that controls the clipping range.

\subsubsection{Advantage Estimation}

To reduce the variance of policy gradient estimates, PPO often employs Generalized Advantage Estimation (GAE). The temporal-difference (TD) residual is computed as:

\[
\delta_t = r_t + \gamma V(s_{t+1}) - V(s_t)
\]

Then, the advantage is estimated via exponentially-weighted TD residuals:

\[
\hat{A}_t = \sum_{l=0}^{\infty} (\gamma \lambda)^l \delta_{t+l}
\]

In practice, GAE is computed using the recursive form:

\[
\hat{A}_t = \delta_t + \gamma \lambda \hat{A}_{t+1}
\]

where \( \lambda \in [0, 1] \) balances the bias-variance trade-off.

\subsubsection{Value and Entropy Loss}

In addition to the clipped surrogate loss for policy updates, PPO includes a squared error loss for value function regression and an entropy bonus to encourage exploration. The total loss is:

\[
L^{\text{PPO}} = L^{\text{CLIP}}(\theta) - c_1 \cdot L^{\text{VF}}(\theta) + c_2 \cdot \mathcal{H}[\pi_\theta]
\]

where \( L^{\text{VF}}(\theta) = \left(V_\theta(s_t) - V^{\text{target}}_t\right)^2 \), and \( \mathcal{H}[\pi_\theta] \) is the policy entropy. The coefficients \( c_1 \) and \( c_2 \) control the weight of each term.

\subsection{Uncontrolled Burgers}
\label{uncontrolled burgers}
\begin{figure}[htbp]
    \centering
    \includegraphics[width=1\linewidth]{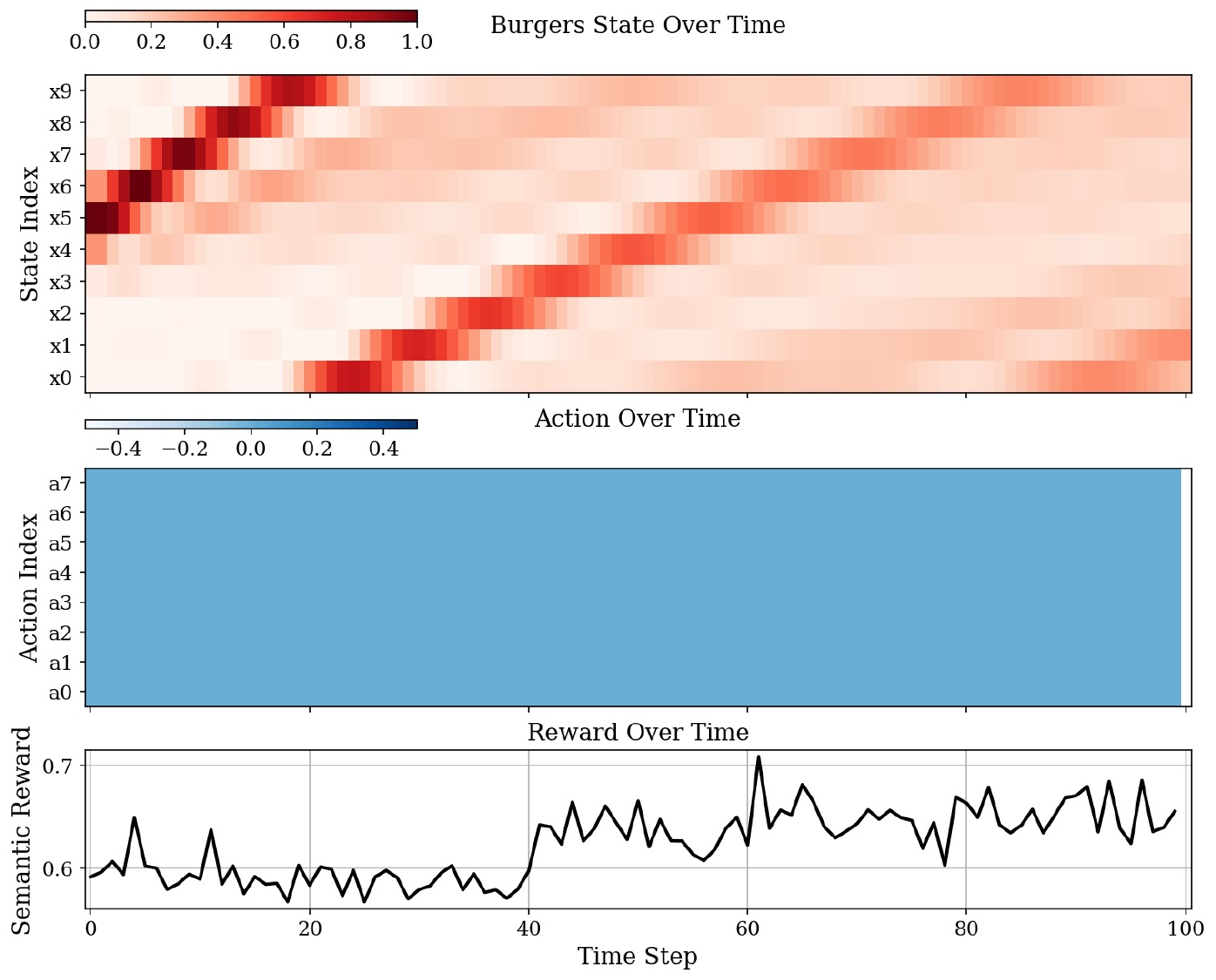}
    \caption{The state evolution in the uncontrolled scenario, where the action is fixed at zero, along with the corresponding reward signal.}
    \label{fig:placeholder}
\end{figure}

\subsection{Geometry of Fluid control}
\label{geometry}

\begin{figure}
    \centering
    \includegraphics[width=1\linewidth]{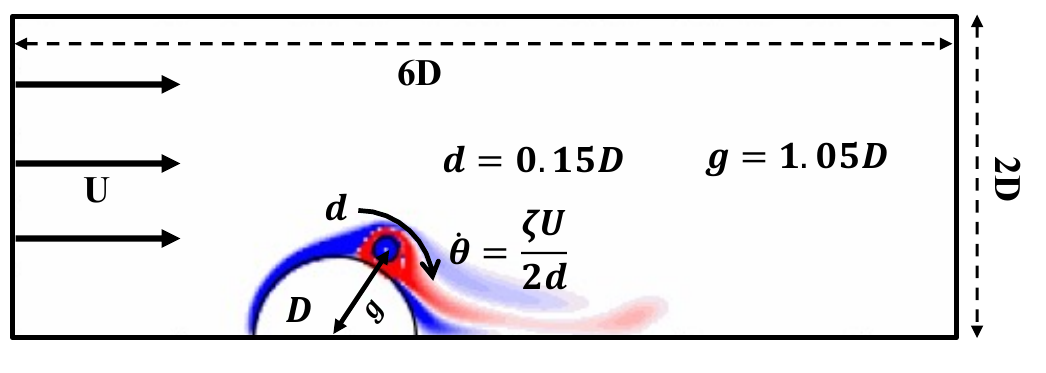}
    \caption{The geometry of the fluid control parameters.}
    \label{fig:geometry}
\end{figure}

\end{document}